\def\arxivversion{1}
\documentclass[letterpaper, 10 pt, conference]{template/ieeeconf}
\IEEEoverridecommandlockouts
\usepackage[pdftex]{graphicx}
\usepackage{amsmath,amssymb}
\usepackage{booktabs}
\usepackage{array}
\usepackage{url}

\usepackage{siunitx} 

\newcommand{\todo}[1]{{#1}}

\newcommand{\MOD}[1]{{#1}}

\newcommand{\todop}[1]{{\{#1\} }}
\newcommand{\comment}[1]{\todop{See comment}}

\newcommand{\method}{\textit{SparseNav}}
\DeclareMathOperator{\ExtractLandmark}{ExtractLandmark}
\DeclareMathOperator{\Seg}{Seg}
\DeclareMathOperator{\Landmark}{Landmark}

\newif\ifarxivversion
\arxivversionfalse
\ifdefined\arxivversion\arxivversiontrue\fi

\title{\LARGE \bf
SparseNav: \MOD{Instruction-conditioned} Sparse Semantic Perception for Training-Free Vision-Language Navigation
}
\ifarxivversion
\author{%
Quanhua Chen, Juhan Kang, Runfeng Lin, and ZiFei Zhang\\
Enquang Feng, Chunran Zheng, Xiwang Dong, and Jiarong Lin\\
{\tt\small Jiarong Lin (corresponding author): ziv.lin@buaa.edu.cn}\\
{\tt\small zivlin@connect.hku.hk}%
\thanks{This work has been submitted to the IEEE for possible publication. Copyright may be transferred without notice, after which this version may no longer be accessible.}%
}
\else
\author{Anonymous Authors}
\fi

\begin{document}
\maketitle
\thispagestyle{empty}
\pagestyle{empty}

\setlength{\abovecaptionskip}{0pt}

\begin{abstract}
Map-based vision-language navigation (VLN) relies on
persistent spatial representations to connect language
understanding with geometric planning. However, acquiring
semantics beyond the needs of the current instruction can
introduce unnecessary perception cost and irrelevant
annotations.
Continuously accumulating unrelated objects may not only
waste computation, but also clutter the visual-spatial
representation consumed by the vision-language model (VLM)
planner. To address this problem, we present \method{},
a training-free framework that follows a \textbf{less-is-more}
principle for semantic navigation.
\method{} persistently maintains a lightweight geometric bird's-eye-view (BEV) map and
sparse landmark memory, acquiring new semantics on demand
using the active sub-instruction to decide what is worth grounding.
An instruction manager first tracks navigation progress
and identifies the active landmark query. An instruction-conditioned
perception mechanism then invokes open-vocabulary segmentation
when the queried landmark is visible and its metric location
can inform the next decision. The resulting landmark memory
supports VLM selection among hybrid frontier and local
directional waypoint candidates. Without any additional training,
\method{} achieves success rates of \SI{42.8}{\percent} on R2R-CE
and \SI{40.7}{\percent} on RxR-CE, both on the Val-Unseen splits.
Controlled ablations examine semantic perception strategies and the
contributions of individual framework components.
Furthermore, we successfully deployed \method{} on a Unitree Go2
quadruped equipped with an Intel RealSense D455 RGB-D camera for
geometric mapping and landmark grounding and a Livox MID-360 LiDAR
for localization, without a prebuilt map. We validated its
effectiveness across multiple indoor environments using
instruction-conditioned waypoint navigation.

\end{abstract}

\section{INTRODUCTION}
Vision-language navigation (VLN) (e.g., \cite{anderson2018,krantz2020}) aims to enable an embodied agent to navigate through an environment according to natural-language instructions. Given a language instruction, VLN requires the agent to jointly perform visual perception, language grounding, spatial reasoning, exploration, progress tracking, and low-level motion execution.
Executing such instructions therefore requires a consistent connection between semantic understanding and geometric planning.

The recent emergence of large vision-language models (VLMs) (e.g.,
\cite{liu2023llava,bai2025qwen25vl,ahn2022,shah2023,huang2023,driesse2023,vint2023})
has substantially improved general visual and linguistic reasoning.
Their open-world knowledge makes them particularly attractive
for training-free VLN, where navigation-specific demonstrations
or task-specific model training can be avoided.
However, directly asking a VLM to predict low-level robot
actions from egocentric images remains difficult.
This difficulty reflects three requirements that
general-purpose visual reasoning alone does not fulfill.
First, image-space reasoning does not by itself provide
a persistent metric representation of traversable space.
Furthermore, long navigation trajectories require
persistent spatial memory that cannot be reliably represented
by an ever-growing sequence of historical RGB images.
Finally, natural-language instructions frequently refer
to concrete landmarks whose approximate visual recognition
is insufficient for precise navigation.

To meet these requirements,
map-based navigation
(e.g., \cite{yokoyama2023,zhang2025,chaplot2020,rosinol2020,peng2023})
provides a natural interface between high-level semantic
reasoning and physical robot motion.
A bird's-eye-view representation explicitly exposes free space,
obstacles, explored regions, robot pose, and candidate destinations.
Recent methods have therefore introduced occupancy maps,
value maps, annotated semantic maps, and hierarchical
semantic-geometric maps into VLM-based navigation.
These representations establish the value of persistent
spatial memory, but leave open how much semantic information
should be acquired for each navigation decision.
We investigate whether maintaining broader semantic coverage
necessarily improves instruction following.

\begin{figure*}[!t]
\vspace{-1.5cm}
\centering
\includegraphics[width=\textwidth,scale=0.5]{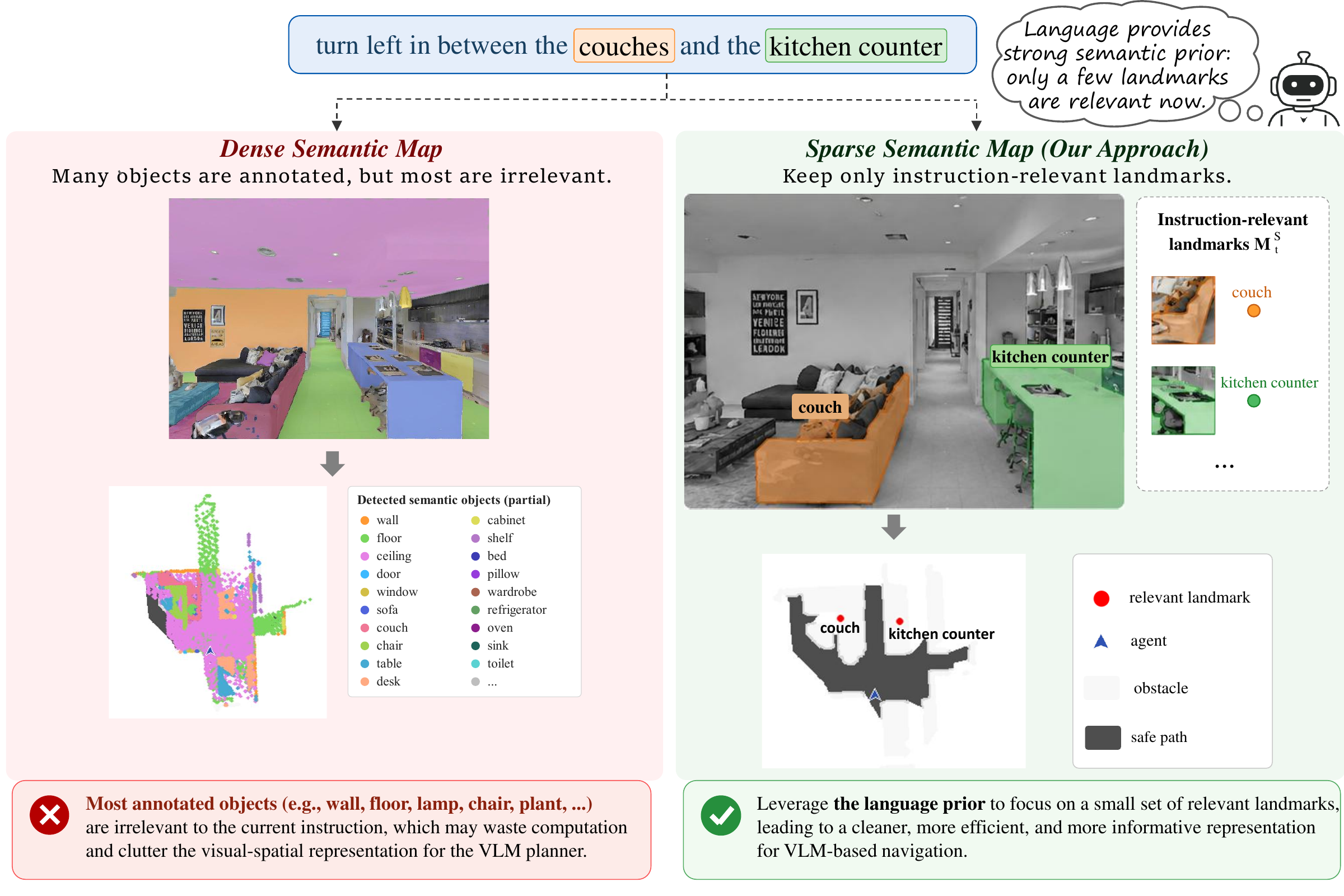}
\vspace{-0.5cm}
\setlength{\abovecaptionskip}{0pt} 
\caption{Language provides a semantic prior for sparse navigation. For the instruction ``turn left in between the couches and the kitchen counter,'' dense annotation includes many objects unrelated to the current decision (left), whereas \method{} grounds the relevant pair while retaining geometry (right). Sparse memory $M_t^S$ stores landmark labels and cluster centers.}
\label{fig:dense-vs-sparse}
\end{figure*}

Instruction-conditioned planning and sub-instruction
tracking~\cite{long2024,chen2025} motivate our hypothesis
that only a small subset of scene semantics needs to be
explicitly grounded for the active navigation objective.
Continuously accumulating unrelated objects may not only
waste computation, but also clutter the visual-spatial
representation consumed by the VLM planner.
For example, consider the instruction fragment in Fig.~\ref{fig:dense-vs-sparse}.
Language already specifies the relevant landmark pair: the \emph{couches} and \emph{kitchen counter}.
Other visible objects need not be semantically annotated to resolve this relation.
Their geometry can still be retained for obstacle avoidance.
Segmenting every chair, lamp, cabinet, painting, and table requires additional computation and may introduce noisy detections and visual-map clutter without necessarily improving the navigation decision.
As illustrated in Fig.~\ref{fig:dense-vs-sparse},
\method{} therefore follows a \textbf{less-is-more}
principle for semantic navigation:
\emph{\textit{Persist geometry, but perceive, ground, and remember semantics only when language makes them relevant.}}

In this work, we instantiate this principle through
an agent-centered, heading-up geometric BEV,
hybrid frontier/local waypoint candidates, and sparse landmark memory.
The active sub-instruction determines the semantic query.
A VLM invokes segmentation only when the queried landmark
is visible and metric grounding can affect waypoint selection.
Mask--depth projection supplies landmark centers for the BEV,
and an instruction manager tracks completed and pending objectives.

Our contributions are summarized as follows:
\begin{itemize}

\item \textbf{\MOD{Instruction-conditioned sparse semantic perception.}}
\MOD{We introduce \todo{a semantic acquisition policy in which the active sub-instruction specifies the landmark queries}\todo{, while a VLM-based trigger determines when metric grounding is needed}.}

\item \textbf{\MOD{A unified spatial interface for training-free VLN.}}
\MOD{We integrate persistent geometry, sparse landmark memory, recent observations, and instruction progress into a BEV-based decision interface. Geometrically generated frontier and local candidates support VLM waypoint selection and classical path planning.}

\item \textbf{Benchmark evaluation and real-robot deployment.}
We evaluate \method{} on R2R-CE and RxR-CE and conduct controlled
ablations of semantic perception strategies and key framework
components. We deployed \method{} on a legged robot and
validated its effectiveness across multiple environments.

\end{itemize}

\section{RELATED WORK}
\subsection{Vision-Language Navigation in Continuous Environments}
VLN was originally formulated using discrete navigation graphs
\cite{anderson2018},
while VLN-CE \cite{krantz2020}
extends the task to continuous environments where an agent
must execute low-level actions rather than select
transitions between predefined viewpoints.
Continuous VLN introduces additional challenges including
collision avoidance, metric localization, waypoint selection,
and control.

Traditional approaches (e.g., \cite{anderson2018,krantz2020,hong2021,mao2021,chen2022})
learn navigation policies from human demonstrations or
simulator-generated trajectories.
While supervised methods have achieved strong benchmark
performance, collecting navigation data is expensive and
learned policies may have difficulty generalizing to
instructions or visual environments that differ substantially
from the training distribution.
These requirements motivate complementary approaches
that reuse pretrained models without navigation-specific training.

Recent work therefore explores training-free or zero-shot
navigation using foundation models.
InstructNav \cite{long2024}
introduces Dynamic Chain-of-Navigation reasoning and
multi-source value maps for generic instruction navigation.
CA-Nav \cite{chen2025}
reformulates navigation as sequential constraint-aware
sub-instruction completion and explicitly maintains
progress constraints.
These methods demonstrate the feasibility of using pretrained
models for navigation without navigation-specific training.

These approaches strongly motivate explicit spatial representations, but the value of persistent geometry does not establish that broader semantic coverage is always beneficial. \method{} follows the training-free paradigm while explicitly controlling which landmarks are grounded and when their observations enter persistent semantic memory. This separates the value of persistent spatial memory from semantic density: geometry remains persistent, whereas semantic information is added only when requested by the active navigation instruction. The resulting map is deliberately incomplete with respect to the scene, and controlled ablations evaluate whether this selective representation preserves navigation quality.

\subsection{Map-Based Memory for VLN}
A persistent spatial representation is particularly valuable for long-horizon navigation because raw observation histories continuously increase in length and provide only implicit spatial relationships.

VLFM constructs occupancy maps from depth observations, identifies exploration frontiers, and combines these frontiers with vision-language semantic values for zero-shot ObjectNav \cite{yokoyama2023}. This work demonstrates the effectiveness of combining geometric exploration with foundation-model semantics.

MapNav introduces an Annotated Semantic Map as a structured memory representation for VLM-based VLN, replacing long histories of raw visual observations with a top-down map containing explicit semantic annotations \cite{zhang2025}.

DreamNav \cite{wang2025dreamnav} combines egocentric view
correction, trajectory-level planning, and imagination-based
prediction to support anticipatory, long-horizon zero-shot
navigation.

\MOD{\method{} retains persistent geometry for traversability reasoning,
while the active sub-instruction determines the landmark queries
and the VLM selectively triggers their metric grounding.
The previously grounded landmarks remain available in sparse memory, \todo{which} makes semantic acquisition an explicit decision, allowing us to study whether instruction-relevant semantics can preserve navigation quality with fewer perception calls.
}

\subsection{Open-Vocabulary Visual Grounding}
Open-vocabulary detection and segmentation \todo{enable the grounding of} natural-language concepts without predefined labels, accommodating varied landmark descriptions in VLN, such as ``the couch next to the window.'' Modern promptable segmentation models such as Segment Anything
support prompt-conditioned mask prediction \cite{kirillov2023}.

SAM 3 \cite{carion2025} extends this capability to concept prompts, including short noun phrases, thereby enabling text-conditioned detection and segmentation.
In our implementation, \method{} uses an open-vocabulary segmentation model as a callable visual perception tool.

\method{} makes segmentation an \textbf{event-driven reasoning tool}: the current navigation state triggers perception only when the VLM identifies a potentially present and task-relevant landmark.

\section{METHOD}
\setlength{\floatsep}{5pt plus 2pt minus 2pt}
\setlength{\textfloatsep}{6pt plus 2pt minus 2pt}
\setlength{\dblfloatsep}{5pt plus 2pt minus 2pt}
\setlength{\dbltextfloatsep}{6pt plus 2pt minus 2pt}
\setlength{\intextsep}{5pt plus 2pt minus 2pt}
\subsection{Problem Formulation}
\begin{figure*}[!t]
\centering
\vspace{-1.8cm}
\includegraphics[width=\textwidth]{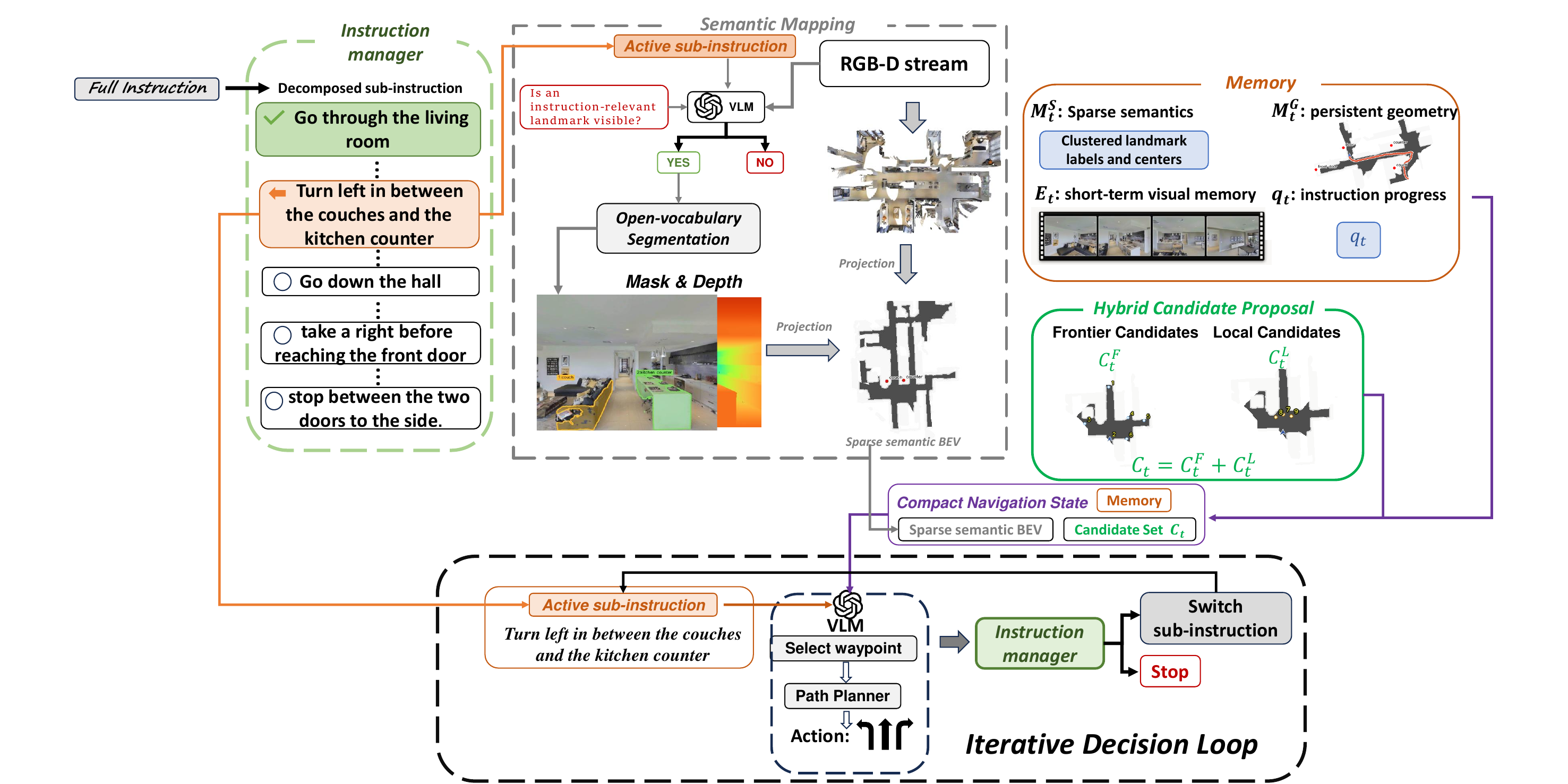}
\vspace{-0.5cm}
\caption{The overview of \method{}. The instruction manager maintains sub-instruction progress and an active landmark query. RGB-D observations and odometry update persistent geometry $M_t^G$; a VLM visibility check gates open-vocabulary segmentation, and mask--depth projection updates landmark labels and cluster centers in $M_t^S$. These memories, recent observations $E_t$, progress $q_t$, and frontier/local candidates form the compact navigation state. The VLM selects a candidate, a path planner generates the trajectory, and execution feedback supports sub-instruction switching or stopping.}
\label{fig:framework}
\end{figure*}

Given instruction $I$ and observations $O\_t$, the agent selects a geometrically valid waypoint $p_t^*\in\mathcal{C}_t$ and executes it with a classical planner. The agent state is
\begin{equation}
\mathcal{M}_t=(M_t^G,M_t^S,E_t,q_t) ,
\end{equation}
where $M_t^G$ is persistent geometry, $M_t^S$ is sparse landmark memory, $E_t$ is recent visual history, and $q_t$ is instruction progress. Fig.~\ref{fig:framework} summarizes the closed loop: the instruction manager supplies the active landmark query, RGB-D observations and odometry update $M_t^G$, and a VLM visibility check gates open-vocabulary segmentation. Mask--depth projection updates labels and cluster centers in $M_t^S$; the VLM then selects from hybrid candidates using $M_t^G$, $M_t^S$, $E_t$, and $q_t$, while planning and execution feedback update subsequent decisions.

\subsection{Geometry-Centric Spatial Memory}
RGB-D or range observations are transformed to a common world frame. Points with $z_{\min}<z<z_{\max}$ are rasterized at resolution $\rho$ to form an obstacle map. Elliptical dilation bridges nearby wall observations, and morphological closing merges fragmented regions and fills small holes:
\begin{equation}
B'_t=B_t\oplus K_{\rm ellipse},\quad
B''_t=(B'_t\oplus K_c)\ominus K_c.
\end{equation}
where \MOD{$B_t$ denotes the binary obstacle map before morphological post-processing.}

Obstacles are inflated by $\lceil r_a/\rho\rceil$ pixels for the robot footprint, using $r_a=0.1\,\mathrm{m}$; free-space accuracy is limited by observed geometry.

For VLM input, the map is translated and rotated to center the robot and align its heading upward:
\begin{equation}
p^{\rm ego}=R(-\theta_t)(p-p_t).
\end{equation}

\MOD{The local frame uses $+x$ forward and $+y$ left. During rasterization, these axes map to image up and image left, respectively, with $u=u_0-y_{\rm ego}/\rho$ and $v=v_0-x_{\rm ego}/\rho$.} Thus image directions match the linguistic notions of forward, left, and right.

\subsection{Hybrid Exploration-Exploitation Waypoint Proposal}
Frontiers discover unknown regions but miss useful waypoints in explored space, such as continuing forward before turning at a sofa.

\method{} therefore defines a hybrid candidate set:
\begin{equation}
\mathcal{C}_t=\mathcal{C}_t^F\cup\mathcal{C}_t^L,
\end{equation}
where $\mathcal{C}_t^F$ contains frontier candidates and $\mathcal{C}_t^L$ local directional candidates. Grounded landmarks guide selection among these two classes.

\subsubsection{Frontier Candidates}
Frontiers are navigable cells adjacent to unknown space; connected cells are clustered, small clusters rejected, and a reachable point with obstacle clearance represents each remaining cluster in $\mathcal{C}_t^F$.

\subsubsection{Local Directional Candidates}
Local proposals preserve motion within explored space. For distance $d_l$ and offsets $\phi_i\in\{-\phi,0,+\phi\}$,
\begin{equation}
l_t^i=p_t+d_l[\cos(\theta_t+\phi_i),\sin(\theta_t+\phi_i)]^\top.
\end{equation}
Candidates in inflated obstacles or disconnected regions are discarded or projected to free space; frontiers support discovery, while local left/front/right proposals support instruction-following maneuvers.

\subsection{Hierarchical Instruction Management}
The instruction manager decomposes $I$ into ordered sub-instructions $S=(s_1,\ldots,s_K)$ and maintains $q_t=(i_t,\mathcal{S}_t^{\rm done})$, where $i_t$ indexes the active item and $\mathcal{S}_t^{\rm done}$ records completed items. The VLM receives completed, current, and upcoming objectives. When the active completion condition is met, the manager records it and advances; progress also determines the active landmark query, avoiding repeated objectives while retaining upcoming context.

\subsection{Instruction-Conditioned Sparse Semantic Perception}
\subsubsection{Task-Relevant Landmark Query}
For active sub-instruction $s_i$, the instruction manager identifies an optional landmark query
\begin{equation}
l_i=\ExtractLandmark(s_{i_t},s_{i_t+1},q_t).
\end{equation}

\todo{This} \MOD{query can contain multiple labels, such as $l_i=\{\text{\textit{couch}},\text{\textit{kitchen counter}}\}$ for a spatial relation. It is derived from the active sub-instruction; at a transition, the immediately upcoming landmark may also be activated when it constrains the next waypoint, without advancing $q_t$.} Perception is restricted to this query rather than all visible objects.

\subsubsection{Semantic Trigger}
The VLM evaluates $(O_t,E_t,s_i,l_i)$ and returns query labels, visibility scores, and a grounding decision $z_t\in\{0,1\}$. Grounding is enabled when the landmark is likely visible and its metric position can affect the next decision; a temporal cooldown suppresses redundant calls unless prior localization is unreliable.

\subsubsection{Open-Vocabulary Landmark Segmentation}
When triggered, the active landmark description is used as the text prompt for an open-vocabulary segmentation model:
\begin{equation}
\mathcal{S}_t=\Seg(O_t,l_i)=\{(\ell_{t,j},S_{t,j})\}_{j=1}^{n_t},
\end{equation}
\MOD{where each $S_{t,j}$ is an independent instance mask with label $\ell_{t,j}$.} We instantiate this step with SAM 3 for text-conditioned concept segmentation \cite{carion2025}.

\subsubsection{2D-to-3D Landmark Projection}
Each mask pixel with valid depth is back-projected using the camera intrinsics $K$ and transformed into the map frame:
\begin{equation}
\MOD{P_{{\rm map},j}^l=T_{{\rm map}\leftarrow{\rm camera}}\Pi^{-1}(S_{t,j},D_t,K),\qquad j=1,\ldots,n_t.}
\end{equation}
After outlier rejection, projected observations are clustered. Sparse semantic memory stores landmark labels and cluster centers observed up to time $t$:
\begin{equation}
M_t^S = \left\{(\ell_j,\boldsymbol{\mu}_j)\right\}_{j=1}^{N_t^S}.
\end{equation}
\todo{where} \MOD{$\ell_j$ is the semantic label and $\boldsymbol{\mu}_j$ the cluster center in the map frame. Each entry is a distinct landmark instance; same-label instances remain separate. Observations update the nearest compatible cluster within the matching threshold or create a new entry.} Previously grounded landmarks remain in memory when no longer queried.

\subsubsection{Landmark-Conditioned Candidate Reasoning}
A landmark center is a semantic reference, not necessarily a traversable target. The VLM combines landmark positions with the active spatial relation to select among collision-checked frontier and local candidates (e.g., couches and a counter define a left-turn passage). The path planner then executes the selected candidate on the geometric map.

\subsection{Visual-Spatial Memory and Waypoint Execution}
The geometric layer stores obstacles, free/unknown space, trajectory, and visited candidates; the sparse layer retains landmark labels and centers. Both are rendered with candidate IDs, while $E_t=\{O_{t-k+1},\ldots,O_t\}$ preserves recent appearance.

The decision input and selected waypoint are
\begin{equation}
X_t=\{M_t^G,M_t^S,E_t,s_{i_t},q_t,\mathcal{C}_t\},\quad
p_t^*=\operatorname{VLM}(X_t)\in\mathcal{C}_t.
\end{equation}
The output identifies a valid candidate and provides a concise rationale. A classical planner computes a collision-free trajectory on the inflated map \cite{hart1968}, separating semantic selection from metric execution.

\section{EXPERIMENTS}
\subsection{Benchmarks}
We evaluate \method{} on the continuous VLN benchmarks \textbf{R2R-CE} and \textbf{RxR-CE}. \textbf{R2R-CE} converts Room-to-Room trajectories into continuous navigation episodes in Habitat environments \cite{anderson2018,krantz2020,savva2019}. \textbf{RxR-CE} contains substantially longer trajectories and richer language descriptions, making it particularly useful for evaluating long-horizon instruction execution \cite{ku2020}. Experiments are conducted on the Val-Unseen splits to assess generalization to previously unseen environments. All experiments, including benchmark evaluation, ablations, and real-robot deployment, use GPT-5 as the VLM \cite{openai2025gpt5}.

Following prior work, we use the standard VLN-CE metrics: Success Rate (SR), Success weighted by Path Length (SPL), Navigation Error (NE), Oracle Success Rate (OSR), and normalized Dynamic Time Warping (nDTW). We report SR, SPL, and NE on both benchmarks, OSR on \textbf{R2R-CE}, and nDTW on \textbf{RxR-CE}.

\subsection{Results on Public Benchmarks}
\begin{table*}[!t]
\vspace{-1.6cm}
\caption{Comparison with supervised and zero-shot VLN methods on Val-Unseen splits. NE is in meters; other metrics are percentages.}
\label{tab:main}
\centering
\small
\setlength{\tabcolsep}{4.3pt}
\renewcommand{\arraystretch}{1.0}
\begin{tabular*}{\textwidth}{@{\extracolsep{\fill}}llrrrrrrrr@{}}
\toprule
Settings & Method & \multicolumn{4}{c}{R2R-CE (Val-Unseen)} & \multicolumn{4}{c}{RxR-CE (Val-Unseen)}\\
\cmidrule(lr){3-6}\cmidrule(lr){7-10}
& & SR$\uparrow$ & SPL$\uparrow$ & NE$\downarrow$ & OSR$\uparrow$ & SR$\uparrow$ & SPL$\uparrow$ & NE$\downarrow$ & nDTW$\uparrow$\\
\midrule
Supervised & SASRA~\cite{sasra_baseline} & 24.0 & 22.0 & 8.32 & -- & -- & -- & -- & --\\
& Seq2Seq~\cite{anderson2018} & \MOD{25.0} & \MOD{22.0} & 7.77 & 37.0 & 13.9 & 11.9 & 12.10 & 30.8\\
& CMA~\cite{anderson2018} & 32.0 & 30.0 & 7.37 & 40.0 & -- & -- & -- & --\\
& NaVid~\cite{zhang2024navid} & 37.4 & 35.9 & 5.47 & 49.1 & 23.8 & 21.2 & 8.41 & --\\
& ETPNav~\cite{an2023etpnav} & 57.0 & 49.0 & 4.71 & 65.0 & 54.8 & 44.9 & 5.64 & 61.9\\
& MapNav~\cite{zhang2025} & 39.7 & 37.2 & 4.93 & 53.0 & 32.6 & 27.7 & 7.62 & 43.5\\
& Dynam3D~\cite{wang2025dynam3d} & 52.9 & 45.7 & 5.34 & 62.1 & -- & -- & -- & --\\
\midrule
Zero-shot & SmartWay$^*$~\cite{shi2025smartway} & 29.0 & 22.5 & 7.01 & 51.0 & -- & -- & -- & --\\
& OpenNav$^*$~\cite{yuan2025opennav} & 19.0 & 16.1 & 6.70 & 23.0 & -- & -- & -- & --\\
& A2Nav$^*$~\cite{a2nav_baseline} & 23.0 & 11.1 & -- & -- & 16.8 & 6.3 & -- & --\\
& InstructNav~\cite{long2024} & 31.0 & 24.0 & 6.89 & -- & -- & -- & -- & --\\
& AO-Planner~\cite{chen2024aoplanner} & 25.5 & 16.6 & 6.95 & 38.3 & 22.4 & 15.1 & 10.75 & 33.1\\
& CA-Nav~\cite{chen2025} & 25.3 & 10.8 & 7.58 & 48.0 & 19.0 & 6.0 & 10.37 & 13.5\\
& DreamNav~\cite{wang2025dreamnav} & 32.8 & 28.9 & 7.06 & 41.0 & -- & -- & -- & --\\
& \textbf{\method{} (Ours)} & \textbf{42.8} & \textbf{35.2} & \textbf{5.96} & \textbf{53.4} & \textbf{40.7} & \textbf{24.1} & \textbf{7.82} & \textbf{48.6}\\
\bottomrule
\end{tabular*}
\par\vspace{0pt}
\begin{minipage}{\textwidth}
\footnotesize
$^*$Partly relies on simulator-labeled training data; zero-shot does not imply that every component is training-free.
Protocols are not necessarily identical across rows.
A dash denotes an unavailable value, not zero. Bold indicates the best performance among zero-shot methods.
\end{minipage}
\end{table*}

\method{} achieves a Success Rate of $42.8\%$ on \textbf{R2R-CE} Val-Unseen, outperforming several earlier zero-shot systems, including InstructNav, CA-Nav, AO-Planner, and DreamNav. On the more challenging \textbf{RxR-CE} benchmark, \method{} achieves an SR of $40.7\%$. Table~\ref{tab:main} also includes supervised methods for context. The reported SR of \method{} is higher than those of NaVid and MapNav on both datasets, while ETPNav achieves a higher SR on both datasets and Dynam3D achieves a higher SR on \textbf{R2R-CE}. These are cross-paper comparisons rather than controlled reruns; differences in evaluated episodes, model backbones, and training data preclude a direct ranking.

Beyond SR, \method{} achieves $35.2\%$ SPL, $5.96\,\mathrm{m}$ NE, and $53.4\%$ OSR on \textbf{R2R-CE}, as well as $24.1\%$ SPL, $7.82\,\mathrm{m}$ NE, and $48.6\%$ nDTW on \textbf{RxR-CE}.

The benchmark results demonstrate the effectiveness of \method{}, but do not isolate the contribution made by sparse semantic perception.

\subsection{Qualitative Analysis}
\begin{figure*}[!t]
\centering
\includegraphics[width=\textwidth]{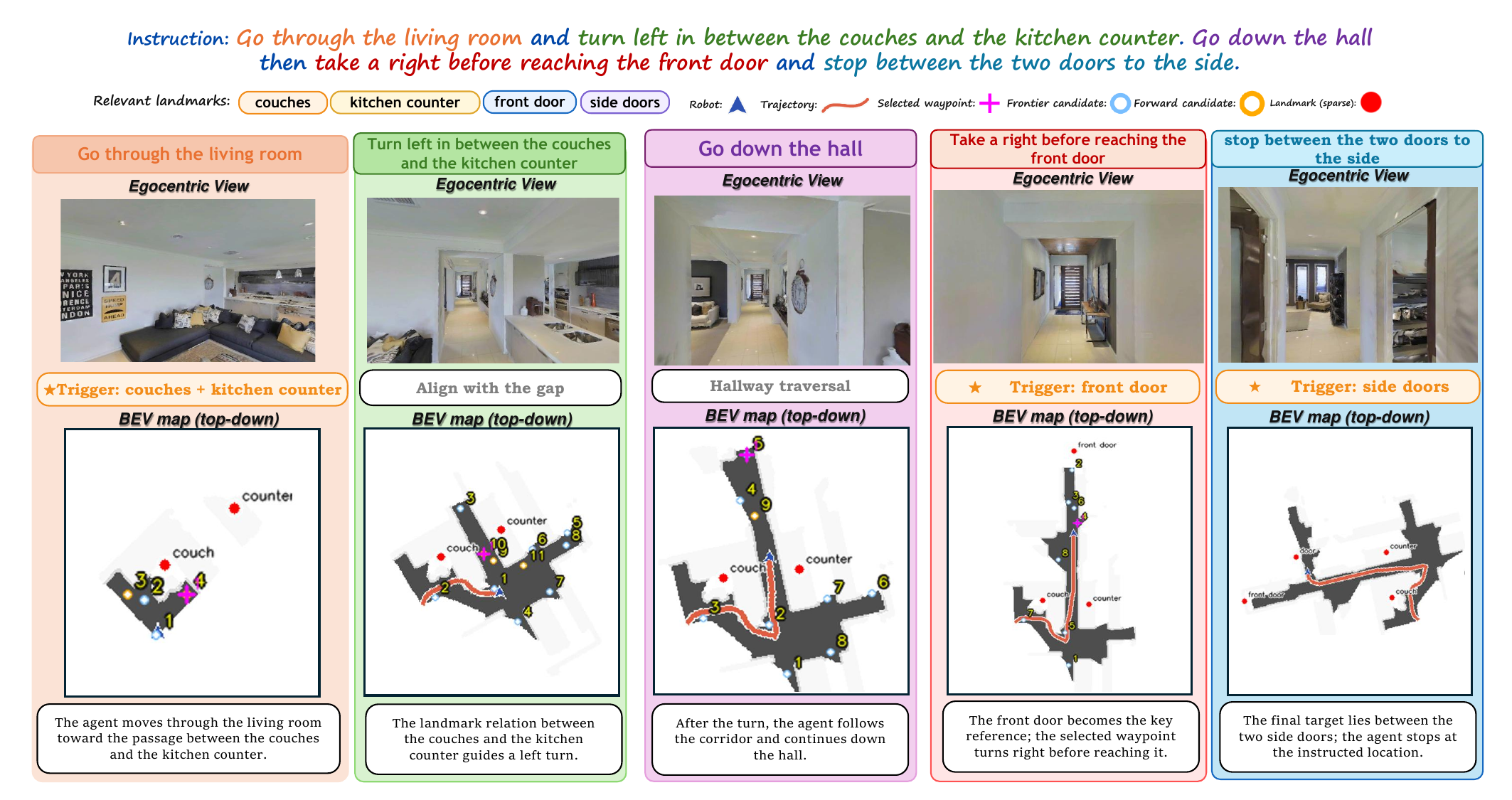}
\caption{Qualitative example of \method{} following the given
instruction. Each stage pairs an egocentric observation with
its corresponding BEV map, illustrating selective landmark
grounding, memory reuse, and waypoint selection.}
\label{fig:qualitative}
\end{figure*}

Fig.~\ref{fig:qualitative} shows a navigation episode in which
\method{} follows the given instruction and stops at the
specified destination.
\MOD{Representative grounding events are highlighted in the figure when instruction-relevant landmarks provide new spatial references for waypoint selection. Other stages proceed using previously grounded landmarks and geometric candidates without a new grounding event being depicted.}

This episode illustrates how \method{} coordinates persistent
geometry with instruction-conditioned semantic perception:
landmark memory supports reuse of acquired spatial evidence,
while new grounding supplies additional evidence as needed.
The agent thus follows the instruction without requiring
semantic acquisition at every sub-instruction.

\subsection{Ablation Studies}

We evaluate semantic perception strategies, waypoint proposals, landmark grounding, instruction memory, semantic triggers, and map orientation through controlled ablations.

\subsubsection{\MOD{Semantic Perception Strategy}}
\MOD{Table~\ref{tab:semantic} separates semantic coverage from perception frequency. The dense-semantic continuous and on-demand settings reach 36.7\% and 38.8\% SR, respectively, whereas restricting queries to instruction-related landmarks gives 40.2\% under continuous perception and 42.8\% with on-demand perception. These controlled comparisons show that both the queried semantic scope and the trigger schedule affect navigation success; the SparseNav configuration combines instruction-related queries with on-demand grounding.}

\begin{table}[t]
\caption{\MOD{Semantic perception strategy.}}
\label{tab:semantic}
\centering
\scriptsize
\begin{tabular}{lr}
\toprule
Semantic strategy & SR$\uparrow$ \\
\midrule
\MOD{No semantic grounding} & \MOD{34.6} \\
\MOD{Dense semantics, continuous perception} & \MOD{36.7} \\
\MOD{Dense semantics, on-demand perception} & \MOD{38.8} \\
\MOD{Instruction-related semantics, continuous perception} & \MOD{40.2} \\
\MOD{\textbf{SparseNav: instruction-related, on-demand perception}} & \MOD{\textbf{42.8}} \\
\bottomrule
\end{tabular}
\end{table}

\subsubsection{Hybrid waypoint proposal}
Table~\ref{tab:waypoint} evaluates the contributions of frontier
and local directional candidates.
Local candidates alone achieve 29.1\% SR, whereas frontier
candidates alone achieve 35.2\%.
Adding a forward local candidate increases SR to 37.4\%,
and including left and right candidates further improves it
to 42.8\%.
These results support combining frontier exploration with
local motion options within explored free space.

\begin{table}[t]
\caption{Waypoint candidate ablation}
\label{tab:waypoint}
\centering
\scriptsize
\begin{tabular}{cccc}
\toprule
Frontier & Local-front & Local-left/right & SR$\uparrow$ \\
\midrule
-- & $\checkmark$ & $\checkmark$ & 29.1 \\
$\checkmark$ & -- & -- & 35.2 \\
$\checkmark$ & $\checkmark$ & -- & 37.4 \\
$\checkmark$ & $\checkmark$ & $\checkmark$ & \textbf{42.8} \\
\bottomrule
\end{tabular}
\end{table}

\begin{figure*}[!t]
\vspace{-1.6cm}
\centering
\includegraphics[width=\textwidth]{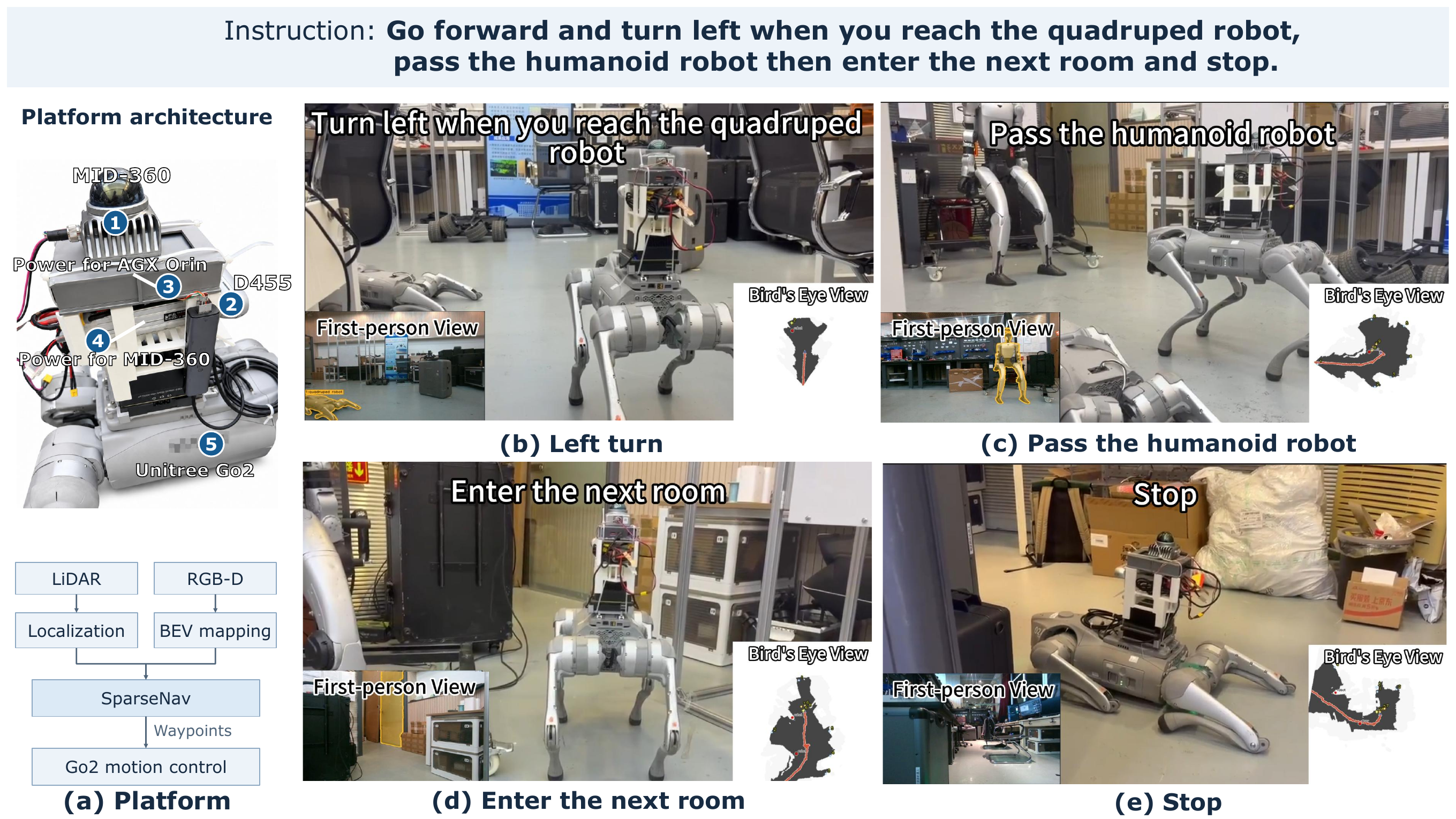}
\caption{Real-robot platform and representative navigation sequence across the evaluated indoor environments: (a) hardware and navigation architecture; (b)--(e) representative execution snapshots showing the left turn, passing the humanoid robot, entering the adjacent room, and stopping.}
\label{fig:real_robot}
\end{figure*}

\subsubsection{\MOD{Landmark Grounding}}
\MOD{Table~\ref{tab:grounding} compares landmark localization strategies while keeping the remaining inputs fixed. Segmentation localization achieves an SR of 42.8\%, compared with 40.2\% for annotation-box localization, and reduces the reported localization error from 0.51 m to 0.12 m. The error is the Euclidean distance between the predicted 3-D cluster center and the true object center. These results suggest that more accurate landmark references can support waypoint selection.}

\begin{table}[t]
\caption{Landmark grounding ablation.}
\label{tab:grounding}
\centering
\scriptsize
\begin{tabular}{lcc}
\toprule
Semantic localization & SR$\uparrow$ & Landmark loc. error$\downarrow$ \\
\midrule
\MOD{No localization} & \MOD{36.5} & -- \\
\MOD{VLM-only localization} & \MOD{34.6} & \MOD{1.56 m} \\
\MOD{Annotation-box localization} & \MOD{40.2} & \MOD{0.51 m} \\
\MOD{Segmentation localization} & \MOD{\textbf{42.8}} & \MOD{\textbf{0.12 m}} \\
\bottomrule
\end{tabular}
\end{table}

\subsubsection{Instruction and Memory}
Table~\ref{tab:memory} evaluates instruction representation and memory on \textbf{RxR-CE}.
With the full instruction, adding temporal memory increases the SR from $29.8\%$ to $31.5\%$. Using sub-instructions further increases the SR to $33.2\%$, and adding progress tracking raises it to $35.0\%$.
\MOD{Adding persistent landmark memory increases SR from 35.0\% to 40.7\%, with instruction decomposition, temporal memory, and progress tracking unchanged.}
These results support the use of temporal context and explicit instruction progress for long-horizon navigation.

\begin{table}[t]
\caption{Instruction and memory ablation on RxR-CE.}
\label{tab:memory}
\centering
\footnotesize
\setlength{\tabcolsep}{2.5pt}
\renewcommand{\arraystretch}{1.00}
\begin{tabular}{@{}lcccc@{}}
\toprule
\shortstack[l]{Instruction\\representation}
& \shortstack{Temporal\\memory}
& \shortstack{Progress\\tracking}
& \shortstack{Persistent landmark\\memory}
& SR$\uparrow$ \\
\midrule
Full instruction
& -- & -- & -- & 29.8 \\

Full instruction
& $\checkmark$ & -- & -- & 31.5 \\

Sub-instruction
& $\checkmark$ & -- & -- & 33.2 \\

Sub-instruction
& $\checkmark$ & $\checkmark$ & -- & 35.0 \\

\MOD{\textbf{\method{} full memory}}
& $\checkmark$
& $\checkmark$
& $\checkmark$
& \MOD{\textbf{40.7}} \\
\bottomrule
\end{tabular}
\end{table}

\subsubsection{\MOD{Semantic Trigger Quality}}
\MOD{Table~\ref{tab:trigger} reports trigger quality on \textbf{R2R-CE}. A trigger is counted as correct when the queried object is visible and relevant to the current instruction; an appearing object followed by a segmentation call is therefore a successful trigger. Sub-instruction-aware triggering achieves the highest precision and recall, at 90.4\% and 99.8\%, respectively.}
For each active sub-instruction $s_i$, the semantic query is $Q_i=\Landmark(s_i)$ and changes as navigation progresses.

\begin{table}[t]
\caption{\MOD{Semantic tool trigger quality on R2R-CE.}}
\label{tab:trigger}
\centering
\scriptsize
\begin{tabular}{lcc}
\toprule
Trigger & Precision$\uparrow$ & Recall$\uparrow$ \\
\midrule
\MOD{Periodic schedule} & \MOD{41.5} & \MOD{45.6} \\
\MOD{VLM visibility only} & \MOD{72.1} & \MOD{99.8} \\
\MOD{Sub-instruction-aware \method{}} & \MOD{\textbf{90.4}} & \MOD{\textbf{99.8}} \\
\bottomrule
\end{tabular}
\end{table}

\subsubsection{\MOD{BEV Coordinate Representation}}
\MOD{Table~\ref{tab:orientation} compares three BEV coordinate representations. The agent-centered but non-rotating configuration reaches 36.2\% SR, while the agent-centered heading-up configuration reaches 42.8\%, compared with 30.3\% for global north-up. This comparison evaluates combined differences in map center and orientation rather than isolating orientation alone.}

\begin{table}[t]
\caption{\MOD{BEV coordinate representation.}}
\label{tab:orientation}
\centering
\scriptsize
\begin{tabular}{lc}
\toprule
Map representation & SR$\uparrow$ \\
\midrule
Global north-up BEV & 30.3 \\
\MOD{Agent-centered, north-up BEV} & \MOD{36.2} \\
\MOD{Agent-centered, heading-up BEV} & \MOD{\textbf{42.8}} \\
\bottomrule
\end{tabular}
\end{table}

\MOD{Taken together, these ablations support the central design of \method{}: persistent geometry provides a spatial basis for navigation, while instruction-related semantic grounding combined with on-demand perception achieves the strongest SR among the evaluated semantic strategies.}
Hybrid waypoint proposals, precise landmark localization, and instruction memory further contribute to navigation performance.

\subsection{Real-Robot Experiment}
\label{sec:real_robot}

\subsubsection{Platform and environment}
\MOD{
Our platform is a Unitree Go2 quadruped equipped with an Intel RealSense D455 RGB-D camera for geometric mapping and landmark grounding, and a Livox MID-360 LiDAR for localization.
We successfully deployed \method{} on this platform without a prebuilt map and validated its effectiveness across multiple indoor environments.}

\subsubsection{Navigation Task and Execution}
Across the evaluated scenarios, instructions define landmark-conditioned
routes. A representative instruction is: \emph{``Go forward and turn
left when you reach the quadruped robot, pass the humanoid robot then
enter the next room and stop.''}
The quadruped robot referenced in the instruction is a scene landmark distinct from the Go2 executing the task.
\method{} tracks sub-instruction progress, selectively grounds relevant landmarks, and retains their positions in sparse semantic memory.
The VLM selects waypoints from the hybrid geometric candidate set, and a path planner generates trajectories for execution.

\subsubsection{Real-World Deployment}
We successfully deployed \method{} in real-world settings and
validated its effectiveness across multiple indoor environments.
Across these environments, instruction progress and grounded landmark
positions guide waypoint selection, while the geometric map supports
path planning. These experiments validate the effectiveness of
integrating instruction-conditioned semantic grounding with geometric
navigation on a real robot.


\section{DISCUSSION AND CONCLUSION}

We presented \method{}, a training-free VLN framework that combines persistent geometric memory with instruction-conditioned sparse semantic grounding.
Geometry represents traversable space, while the active instruction determines which landmarks require grounding.
Hybrid frontier and local waypoint candidates, accumulated landmark memory, and explicit instruction progress jointly support VLM-based navigation.

\method{} achieves success rates of 42.8\% on R2R-CE and 40.7\% on RxR-CE, both on the Val-Unseen splits.
\MOD{Ablation studies show that instruction-related, on-demand semantic grounding achieves higher SR than the evaluated dense-semantic settings.}
Additional ablations support the contributions of hybrid waypoint proposals, precise landmark localization, instruction memory, and agent-centered map orientation.
Successful deployment on a Unitree Go2 across multiple indoor environments further validates the effectiveness of the framework.

Together, these findings suggest that task-relevant sparse semantic grounding can be sufficient for semantic navigation: persistent spatial geometry can support effective instruction following without requiring a semantically complete scene map.
The results highlight the potential value of selectively acquiring and reusing task-relevant landmarks.
However, they do not establish that reduced visual clutter alone causes the performance gain, since selective grounding also changes the semantic evidence available to the planner.

\section{LIMITATIONS AND FUTURE WORK}

Despite its effectiveness, \method{} has several limitations.
First, its rule-based waypoint generator restricts the VLM to a predefined set of frontier and local directional candidates.
Although these candidates support both exploration and local maneuvers, they may omit waypoints that better match the instructed route, thereby limiting the flexibility of navigation.
Second, selective grounding reduces segmentation overhead, but VLM inference remains a major computational bottleneck.
In a sequential perception--reasoning--execution pipeline, waiting for waypoint decisions can delay robot motion and increase task completion time.
Third, although \method{} achieves competitive zero-shot results, its reported success rates on R2R-CE and RxR-CE remain below those of some navigation-trained methods.
Differences in training data and evaluation settings prevent attributing this gap solely to the absence of navigation-specific training.

Future work will investigate learned waypoint proposals to provide more flexible, instruction-conditioned candidates while retaining geometric feasibility checks.
Such extensions would be evaluated separately from the current training-free setting.
We will also explore more efficient VLM inference and asynchronous planning and execution, allowing the robot to follow a validated local trajectory while preparing the next decision.
This design will require mechanisms to detect stale plans and respond to changes in the observed environment.
Finally, further real-world evaluations across more diverse layouts, landmark configurations, and instructions will assess the robustness and generalizability of the framework.

\bibliographystyle{template/IEEEtran}
{\let\footnotesize\scriptsize\let\small\scriptsize\bibliography{references}}

\end{document}